\documentclass[10pt,twocolumn,letterpaper]{article}

\usepackage{cvpr}
\usepackage{times}
\usepackage{epsfig}
\usepackage{graphicx}
\usepackage{amsmath}
\usepackage{amssymb}
\usepackage{framed,multirow}

\usepackage[breaklinks=true,bookmarks=false]{hyperref}

\cvprfinalcopy 

\def\cvprPaperID{****} 

\begin{document}

\title{Single Image Action Recognition using Semantic Body Part Actions}

\author{Zhichen Zhao$^1$, Huimin Ma$^1$, Shaodi You$^2$\\
$^1$Department of Electronic Engineering, Tsinghua University\\
$^2$Data61, CSIRO, Australia (NICTA)\\
$^1${\tt\small \{zhaozc14@mails., mhmpub@\}tsinghua.edu.cn},
$^2${\tt\small Shaodi.You@data61.csiro.au}
}

\maketitle

\begin{abstract}
   In this paper, we propose a novel single image action recognition algorithm which is based on the idea of semantic body part actions.
   Unlike existing bottom up methods, we argue that the human action is a combination of meaningful body part actions. 
   In detail, we divide human body into five parts: head, torso, arms, hands and legs. And for each of the body parts, we define several semantic body part actions, \eg, hand holding, hand waving. These semantic body part actions are strongly related to the body actions, \eg, writing, and jogging.
   Based on the idea, we propose a deep neural network based system: first, body parts are localized by a Semi-FCN network. Second, for each body parts, a Part Action Res-Net is used to predict semantic body part actions. And finally, we use SVM to fuse the body part actions and predict the entire body action.
   Experiments on two dataset: PASCAL VOC 2012 and Stanford-40 report mAP improvement from the state-of-the-art by 3.8\% and 2.6\% respectively.
\end{abstract}

\section{Introduction}
\begin{figure}[!t]
\includegraphics[width=1\linewidth]{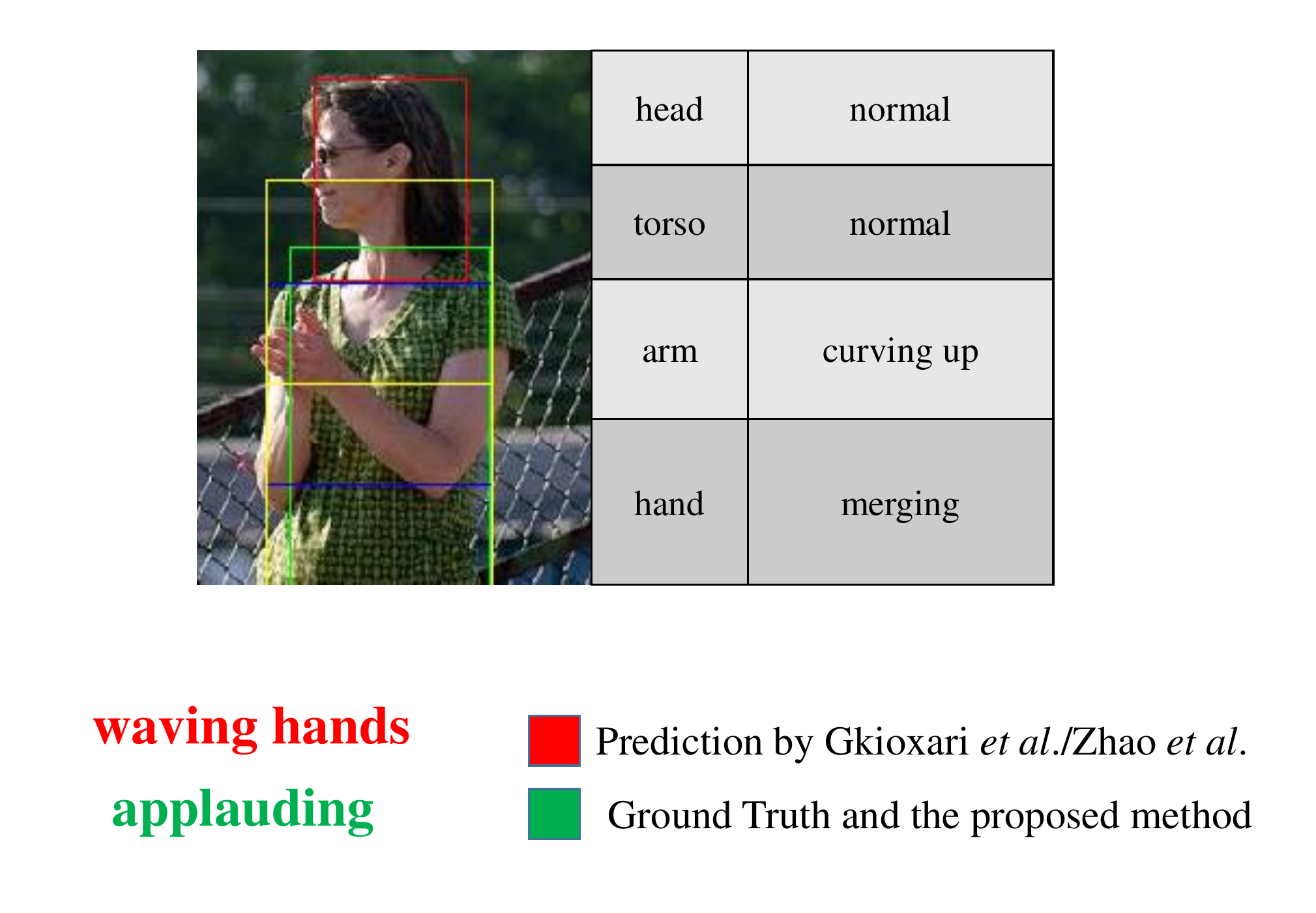}
\caption{Inferring the semantic action by semantic body part actions. A person ``applauding" is classified as ``waving hands"  by the state-of-art methods~\cite{contextualaction,topdown}, because the hands are away from body which follows the pattern for waving. Our method, however, makes the correct prediction by noticing the semantic body part action that the hands are ``merging".
}
\label{insp}
\end{figure}

\begin{figure*}[!t]
\centering
\includegraphics[width=1\linewidth]{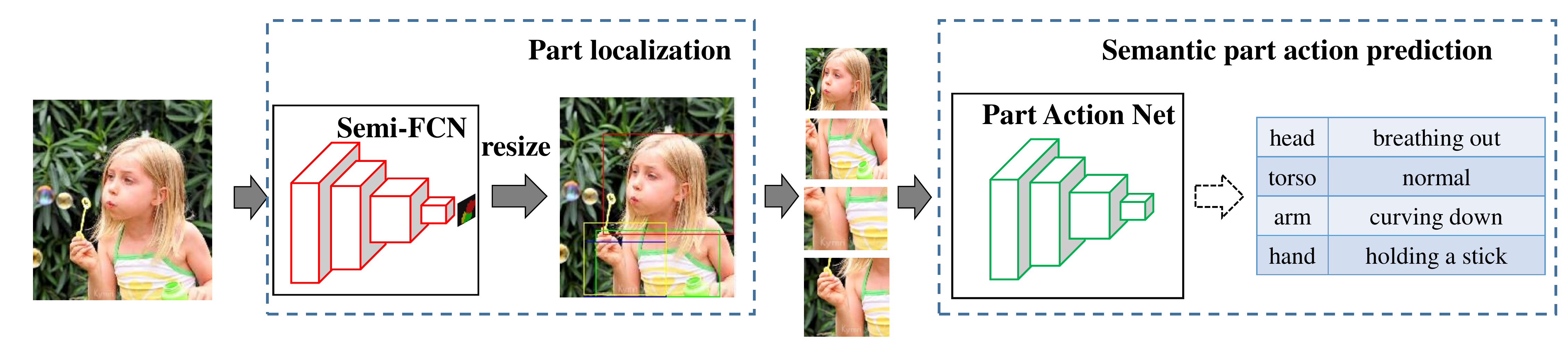}
\caption{The proposed framework for semantic body part action prediction. 
}
\label{framework}
\end{figure*}

Single image action recognition is a high level computer vision task which aims to identify the human action in still image where location prior is provided.
It is a core task in computer vision which enables better performances of image caption \cite{cap1}, image and video analysis \cite{c3d}, human-computer interactions \cite{hoi} and \etc.

Early single image action recognition methods exploit bottom up solutions with raw body parts spatial relations~\cite{Yao11,SMSP,poselet}. 
Recently, benefited from deep neutral networks, body part recognition and object detection perform much better, which eventually significantly improves the performance of body action recognition~\cite{wholeandpart,contextualaction,topdown}. 

However, we argue that human action is not only combination of raw body parts and spatial relations. Rather than that, body is actually presented by semantic local part actions.
For example, as shown in Fig.\ref{insp}, here the hands of the persons are away from the body, and is thus classified as ``waving hands" by Gkioxari \etal and Zhao \etal ~\cite{contextualaction,topdown}. 
However, by noticing that her hands are clapping with each other, one might infer that she is actually applauding. 
In our method, such semantic hand action is labeled as ``hands:merging". And our method makes the correct prediction using such semantic body part actions.

In this paper, we propose a novel single image action recognition algorithm which is based on the idea of semantic body part actions. 
As illustrated in Fig.\ref{framework}, first, we locate five body parts (head, torso, arms, hands and legs) using a ``semi"-Fully Convolutional Network (semi-FCN).
Second, and most importantly, we predict the semantic body part actions. In detail, we define a set of semantic body part actions, \eg,``head: looking up", ``hand: supporting" and ``leg: crouching". We also provide a training dataset with body part annotations as our defined categories. Based on these data, we train the part action prediction Res-Net.
Finally, we link the body part actions to the entire body action: we select part features by Linear Discriminant Analysis (LDA), and concatenate the image feature and selected part features; and use a SVM to determine the final body action.

We evaluate our method on two popular but challenging dataset: (1) PASCAL VOC 2012 \cite{pascalvoc2012} and (2) Stanford-40 \cite{Yao12}. Our method reports improvements from the state-of-the-art by 3.8\% and 2.6\% (mean average precision, mAP) respectively. 


The contributions of this paper are:
first, we propose concept that human action can be inferred by local part actions, which is a intermediate level semantic concept.
Second, we propose the methodology which combines body actions and part actions for action recognition. 
And finally, the proposed method achieves a state-of-the-art performance.




The rest of the paper is organized as follows:
Sec.\ref{related} introduces relate work on action recognition.
In Sec.\ref{design}, we introduce the idea of semantic body part actions. Sec.\ref{main} introduces the full system, followed by experiments in Sec.\ref{experiments}. Sec.\ref{conclusion} is the conclusion.
 
\begin{figure*}[!t]
\centering
\includegraphics[width=1\linewidth]{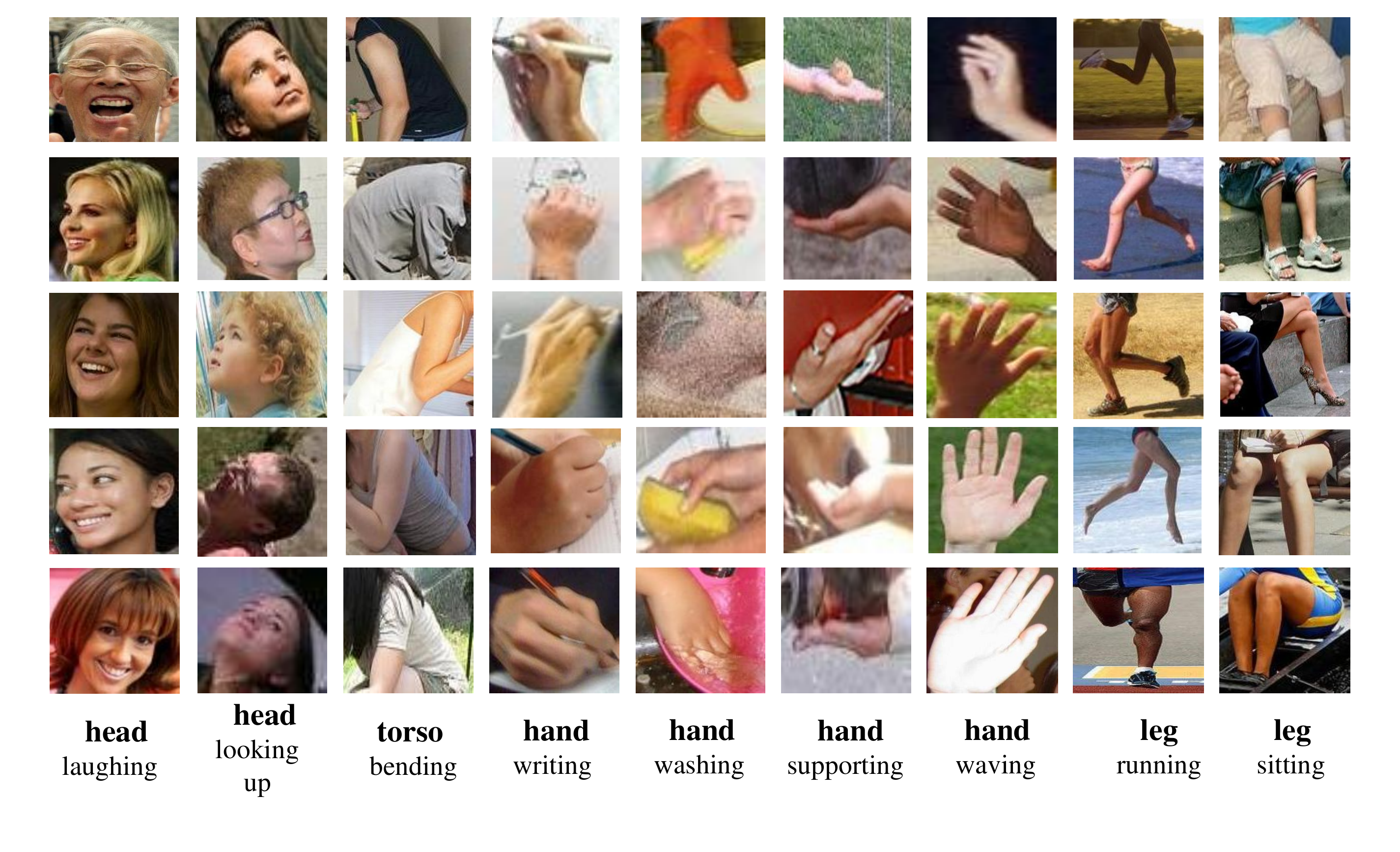}
\caption{Examples of semantic body part actions. Images are from Stanford-40 dataset \cite{Yao12}.
}
\label{partaction}
\end{figure*}

\begin{figure*}[!t]
\centering
\includegraphics[width=1\linewidth]{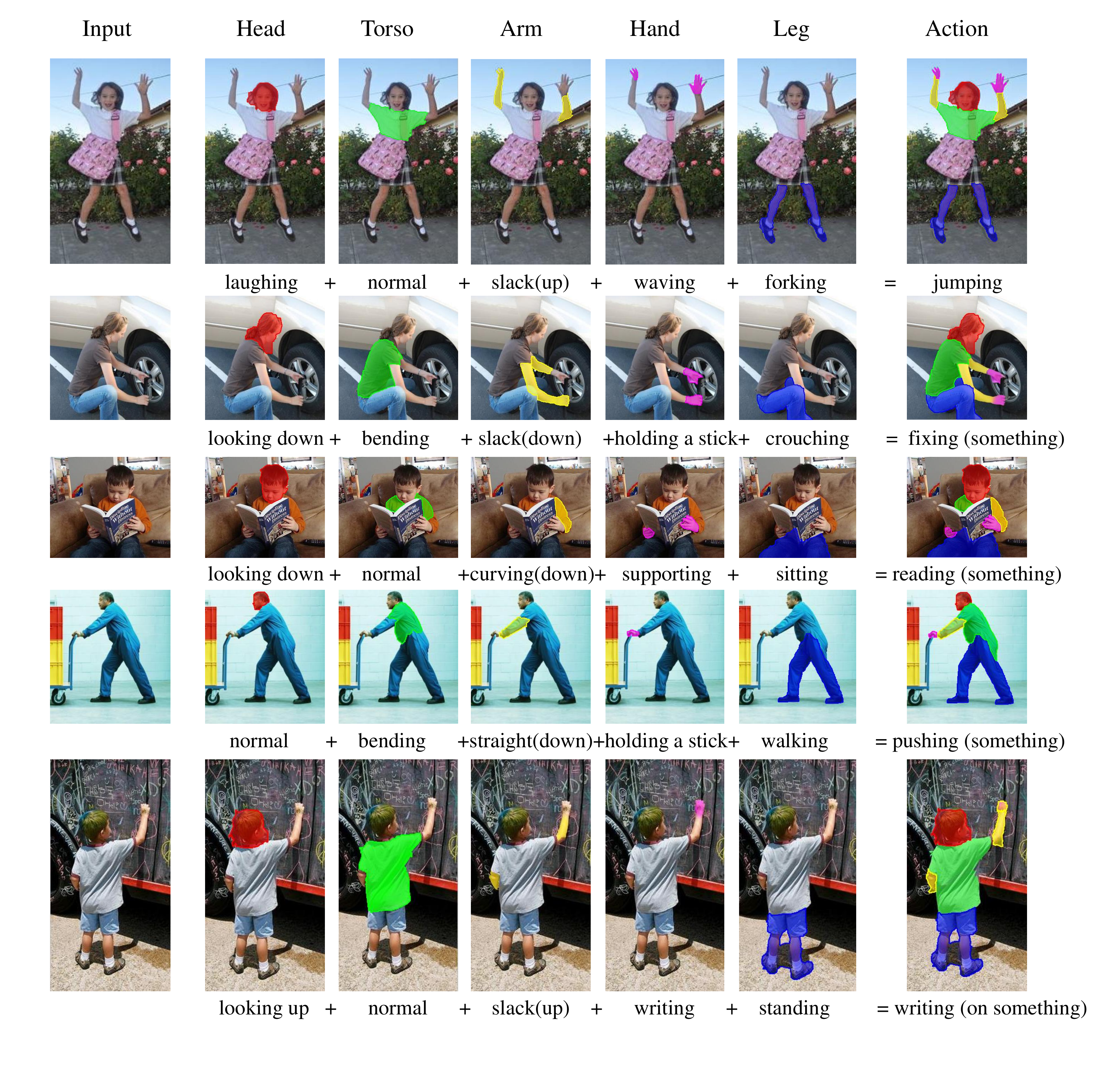}
\caption{Entire body actions as combination of semantic body part actions.
}
\label{mask}
\end{figure*}

\section{Related work}\label{related}
\noindent \textbf{Single image action recognition.} 
There are mainly three existing strategies for single image action recognition: context-based approaches, part-based approaches and template-based approaches. For context-based approaches, cues from interactive objects are critical. Gkioxari \etal \cite{contextualaction} employ object 
proposals~\cite{selectivesearch} to find proper interactive objects.
Zhang \etal \cite{mini} propose a method that segments out the precise
regions of underlying human–object interactions with minimum annotation efforts.

Template-based approaches focus on action structures. Desai and Ramanan~\cite{phrasetree} learn a tree structure for each action, treating poses and interactive objects as leaf nodes and modeling their relations. 
Yao and Li~\cite{25D} combine view-independent pose information and appearance information, and propose a 2.5D representation.

The human body parts provide rich information for action. For action recognition and fine-grained recognition, part-based methods have shown promising results~\cite{parteccv,onevsall,wholeandpart}. A typical approach to combine global cues and body part cues is concatenating their features~\cite{SMSP,wholeandpart}. 
Based on such framework, most existing part-based methods assign parts by the same labels with the corresponding categories. However, we argue that for action recognition, part action does not belong to a certain category. One kind of part action can exist in many similar actions and we can consider part actions independently. To distinguish part actions, we define a set of part action categories, and train a specific part action classification network to extract discriminative features.

\noindent \textbf{Fully convolutional network.}
Different from normal deep CNNs, fully convolutional networks (FCNs) drop all fully connected (fc) layers, add deconvolutional (deconv) layers and product pixel-wise prediction. It is first introduced to semantic segmentation~\cite{fcn}. Recently, some method~\cite{finefcn} employs FCN framework to localize parts in part-based approaches. 

\section{Semantic Body Part Actions}\label{design}

\begin{table}[tbp]
\caption{\label{subcat}List of part actions.}
\centering
\begin{tabular}{|c|c|c|c|}
\hline
\multirow{11}{*}{head} & breathing out & \multirow{15}{*}{hand} & cutting\\
 & drinking &&merging\\
 & laughing&&printing\\
 & looking up&&proping\\
 & looking down&&slack\\
 & looking through&&supporting\\
 & normal&&washing\\
 & speaking&&waving\\
 & tooth brushing&&writing\\
 & sucking&&holding a bottle\\
 \cline{1-2}
 \multirow{4}{*}{torso}&bending&&holding a stick\\
  
 & fading away&&holding a phone\\
 & normal&&holding a camera\\
 & lying & & holding a cigarette\\
\hline
 \multirow{6}{*}{arm} & curving ip&\multirow{6}{*}{leg}&crouching\\
  & curving (down)&&forking\\
  & slack (up)&&running\\
  & slack (down)&&sitting\\
  & straight (up)&&standing\\
  & straight (down)&&walking\\
  \hline
\end{tabular}
\end{table}


We argue that the entire human body action is not only a direct combination of body parts, but there exists a mid-level semantic, local part actions. And they have highly relate to entire body actions. For example, ``cooking" are strongly related to part action ``hand: cutting".

\subsection{From body parts to semantic part actions}
The part action is a mid-level semantic.
Extreme detailed body part segmentation and localization is impractical for single image action recognition. Also, the entire body action is a semantic rather than geometry. Therefore, it is not necessary to have extremely detailed body part definition but rather a set of semantic body parts.
Specifically, we define five body parts: head, torso, arms, legs and hands.
And we define their semantic actions. As illustrated in Fig.\ref{partaction}, each body part has a relative simple action. For example the head can be ``laughing'', ``looking through'', ``looking up'' and \etc.


A full list of possible actions is provided in Tab.\ref{subcat}. 
In detail: when define part actions, we consider both action diversity and compactness: final combinations should cover as many as possible of different human actions, and each kind of action should contain as few subcategories as possible.
For head, we define $10$ kinds of actions: ``breathing out", ``brushing teeth", ``drinking",``laughing", ``looking down", ``looking up", ``looking through", ``speaking",``sucking" and ``normal".
Similarly, for torso we define ``bending", ``fading away" ``lying" and ``normal". 
For arms we define six types of actions, combined from three states (``slack", ``curving" and ``straight") and two kinds of upper arm orientation(``up" and ``down"). 
For legs, we define ``crouching", ``forking", ``running", ``standing", ``walking" and ``sitting".
Actions of hands are versatile and critical for the final action, we define $14$ categories. 
Other than general actions like ``supporting", ``waving", ``slack", we also define subtle actions such as ``printing", ``holding a bottle", ``washing" and ``cutting". 
The entire part action set contains $40$ categories. 
Fig.\ref{partaction} shows some examples. 

Because there are no such labeling off-the-shelf, we collect annotations from the training set of Stanford-40 which are manually labeled by volunteers. The full annotation is provided in the supplementary materials.

\subsection{From semantic part actions to entire body actions}

The high level semantic actions can be inferred by semantic body part actions.
For example, if semantic body part actions are ``head:looking down", ``torso:bending", ``arms:slack down", ``hands:holding a stick" and ``legs:crouching'', even without seeing the image, we can guess the entire action is ``fixing something". 
In Fig.\ref{mask} we show more examples on the high level semantic are formulated by a combination of mid-level semantics.


\section{Action Recognition System}\label{main}
In this section, we introduce semantic part action prediction system and the entire body action prediction system. 

As illustrated in Fig.\ref{framework}, first, Semi-FCN is used to localize body parts. Second, a part action prediction module is implemented using Res-Net.
Then a SVM module is used to predict final action. We introduce the three modules correspondingly in the following subsections.

\subsection{Body part localization}\label{loc}

We employ Fully Convolutional Network (FCN) to efficiently localize multiple body parts.
Unlike previous methods~\cite{fcn,finefcn}, which relies on a fine estimation. Our method only requires part bounding boxes, which significantly improves the efficiency and robustness.
We propose a `semi-FCN" structure: unlike the existing method, we remove all the up-sample layers and resize the ground truth annotations as the same size with the conv5 feature maps. 
We add a loss layer after the conv5 layer, and still calculates segmentation loss. 
Specifically, A VGG-16 based FCN~\cite{fcn} is employed. We resize the input image to $512\times 512$, the conv5 feature map has a size of $16\times 16$. The ground truth labeling are also resized to $16\times 16$, as shown in Fig.\ref{framework}. 

The reasons why we remove deconv layers and obtain only low-resolution prediction are two-fold: (1) Bounding boxes are sufficient to capture body parts for our purpose.
(2) High-resolution ground truth requires intensive human labor for labeling. On the contrary the low-resolution bounding box can be generated more efficiently. 
Actually, the part bounding box can be automatically generated based on existing human joint labeling from pose estimation dataset.

\paragraph{Dataset and Training} 
To generate labels for training and testing, we use two dataset: PASCAL Part dataset~\cite{pascalpart} and MPII~\cite{andriluka14cvpr}. PASCAL-Part dataset contains a set of additional labeling for PASCAL VOC 2010~\cite{pascalvoc2010}. It defines part on all of the $20$ categories and provides pixel-wise segmentation ground truth for each part.
The training and validation set contain $10,103$ images, from which we only use images that contain people.
The MPII dataset contains $25$K images and more than $40$K instances. It provides center coordinates for each joint, which can be used to generate low-resolution ground truth. 
In this paper we use both datasets. For PASCAL-Part dataset we resize the ground truth into $16\times 16$ grid by the nearest-neighbor algorithm. For MPII we label each pixel as the same category with its nearest joint, obtaining approximate pixel-wise maps, and then resize.

We train our semi-FCN by SGD with momentum for 60 epochs. We use a minibatch size of 50 images and fix learning rates as $0.001$. For $16\times 16$ segmentation, our semi-FCN reaches 65.9\%  pixel accuracy. Fig.\ref{vis} visualizes final bounding boxes. Our semi-FCN sometimes mislabels part edges, but it always predict the locations correctly. Note sometimes the network may fail to localize small parts (\eg hands), if its father node (arms) can be localized, we localize this part on the endpoint of the father node (for hands we take two endpoint, and average pool features of these two regions), otherwise we use prior mean locations on the training set.

\subsection{Semantic part action prediction}
The part action prediction network is implemented by a ResNet-50~\cite{He2015}. 
And it is trained using our dataset. The dataset is previously introduced in Sec.\ref{design}.
In detail the ResNet is trained on the ImageNet \cite{imagenet} as initialization. 
After that, to train our part action prediction network. The learning rate is set to $10^{-6}$. After $50$K iterations the network reaches accuracy of 60.2\% on our own training and validation sets.

\subsection{Entire body action detection}
As we have argued, a specific entire body action is highly related to one of more semantic body part actions. That is to say, on one hand, some part actions (or combination) will be related to one specific entire body action. On the other hand, some part actions might be irrelevant to a specific entire body action.
Therefore, We use a Linear Discriminant Analysis (LDA) to learn the connections between different part actions and the final body actions.

In detail, for a given part $\mathbf{x}^{(k)}$, we denote its part action feature as $\mathbf{f}^{(k)}$, $k=1,2,\dots$, where $k$ is index of defined parts. For a given human action category $c$, 
we measure if a part action is discriminative for this body action by Linear Discriminant Analysis (LDA): 
for part $k$, the within-class scatter matrix can be written as
\begin{equation}
S_w^{(k)}=\sum^{C}_{c=1}\sum_{\mathbf{x}^{(k)}\in c}^{N}(\mathbf{f}_i^{(k)}-\mathbf{u}_c^{(k)})(\mathbf{f}_i^{(k)}-\mathbf{u}_c^{(k)})^T,
\end{equation}
where $i$ is index of sample and $\mathbf{u}_c$ is mean of this category. 
Similarly, the between-class scatter matrix can be written as
\begin{equation}
S_b^{(k)}=\sum^{C}_{c=1}n_c(\mathbf{u}^{(k)}-\mathbf{u}_c^{(k)})(\mathbf{u}^{(k)}-\mathbf{u}_c^{(k)})^T,
\end{equation}
where $\mathbf{u}$ is the global mean, and $n_c$ is sample amount of category $c$. A discrminative part always yields large between-class variance and small within-class variance, \ie, a large $J(\mathbf{w})$:
\begin{equation}
J^{(k)}(\mathbf{w})=\frac{\mathbf{w}^TS_b^{(k)}\mathbf{w}}{\mathbf{w}^TS_w^{(k)}\mathbf{w}}.
\label{LDA}
\end{equation}
The maximum of $J^{(k)}(\mathbf{w})$ measures the discrimination of part $k$. For each action, we optimize the maximum of $J^{(k)}(\mathbf{w})$, and rank parts by the corresponding scores. 
We choose top $M$ parts as selected discriminative parts and use them to predict the final body action.
Specifically, in this paper we define up to $5$ type of parts, and set $M=2$.

\paragraph{Full implementation}
As a common practices \cite{SMSP,wholeandpart}, and to optimize the final performance, we also incorporate 
``body action classification network".
We combine both networks to extract part features. Denote features extracted by the body action network and the part action network by $\mathbf{f}_{b}$ and $\mathbf{f}_{p}$ respectively, we can combine them as body\&part representations for part $k$:
\begin{equation}
\mathbf{f^{(k)}}=[\mathbf{f}_b^{(k)T},\mathbf{f}_p^{(k)T}]^T. 
\end{equation}
The final representations are combinations of bounding box image feature and selected part features:
\begin{equation}
\mathbf{f}=[\mathbf{f}_{bbox}^T,\mathbf{f}^{(1)T},\mathbf{f}^{(2)T},\dots \mathbf{f}^{(M)T}]^T. 
\end{equation}

We emphasize semantic body part feature more than the entire image feature, that we weight each of the part features by $0.5$, which is 2.5 in total and the entire image features by $1$: 
\begin{equation}
\mathbf{f}=[\mathbf{f}_{bbox}^T,0.5\mathbf{f}^{(1)T},0.5\mathbf{f}^{(2)T},\dots 0.5\mathbf{f}^{(M)T}]^T. 
\end{equation}
A linear SVM is used for the final prediction.

To fully justify the proposed semantic part prediction, in the experiments, we report the decomposed results in Tab.\ref{verify}.

\begin{table}[tbp]

\caption{Decomposed Performance (mAP) of using different combination of features. 
In this table, $N_b$ denotes the body action network and $N_p$ denotes the part action network.
}
\centering
\begin{tabular}{|c|c|c|}
\hline
method & PASCAL VOC (val) & Stanford-40\\
\hline
no parts & 81.4 & 80.2\\
\hline
part+$\mathbf{N}_b$  & 82.0 & 81.2\\
\hline
part+$\mathbf{N}_p$ & 81.8 & 81.0\\
\hline
part+$\mathbf{N}_b$+$\mathbf{N}_p$ & 83.2 & 82.4\\
\hline
part+$\mathbf{N}_b$+$\mathbf{N}_p$ &\multirow{2}{*}{84.0} &\multirow{2}{*}{83.4}\\
(part selected) &&\\
\hline

\end{tabular}\label{verify}
\end{table}

\section{Experiments}\label{experiments}
We conduct intensive experiments to validate the proposed method. The results show that our method reaches superior results compared with the state-of-the-art methods. Especially, on PASCAL VOC 2012 dataset, our performance is 3.8\% better than the state-of-the-art and Standford-40 is 2.6\% better.

\subsection{Experimental setup}
\paragraph{Network.} In this paper we have 3 networks: semi-FCN, body action classification network and body action classification network. For semi-FCN, we use VGG-16~\cite{vgg} as basic network. Body action network and part action network are both ResNet-50~\cite{He2015}. We have provided parameter settings for semi-FCN and part action network in Sec.\ref{main}. For body action classification network, we set the learning rate to be $10^{-6}$ and the mini-batchsize to be $30$. All $3$ networks are trained on a single Titan X GPU.

\paragraph{Dataset.} As common practice, we use two challenging datasets: 1) PASCAL VOC 2012~\cite{pascalvoc2012} and 2) Stanford-40~\cite{Yao12}.
The PASCAL VOC dataset contains 10 different actions. 
For each of the action type, 400-500 images are used for training and validation, and the rest are used for test.
The Stanford-40 dataset contains 40 actions and uses 100 images for training. 
Since both dataset have only limited amount of training images, we augment the training set by flipping and cropping.

\paragraph{Metrics.} The performance is evaluated based on average precision (AP). All of the measure functions are provided in VLFeat library~\cite{VlFeat}.

\subsection{Results and analysis}
To measure the effectiveness of our approach, we implement experiments under different conditions. Tab.\ref{verify} shows some comparison. In Tab.\ref{verify}, we denote body action network by $\mathbf{N}_b$ and part action network by $\mathbf{N}_p$. On both dataset, using part information is critical. Body action network outperforms part action network. However, since two networks are trained to learn different semantics, combining them improves the final performance.

Notice that $\mathbf{N}_p$ is only trained on Stanford-40 dataset but help improve the performance on PASCAL VOC 2012 dataset, which implies that independent part actions can be easily applied for other custom dataset, without re-annotation.

\begin{table*}[tbp]
\caption{\label{vocval}Preformance (mAP) on the PASCAL VOC 2012 Action validation set}
\footnotesize
\centering
\begin{tabular}{|c|cccccccccc|c|}
\hline
\multirow{2}{*}{method}  & \multirow{2}{*}{jumping}&\multirow{2}{*}{phoning} &
playing &\multirow{2}{*}{reading}&
riding&riding&\multirow{2}{*}{running}&
taking&using&\multirow{2}{*}{walking}&\multirow{2}{*}{mAP}\\
&&&instrument&&bike&horse&&photo&computer&&\\
 \hline
 Whole\&Parts~\cite{wholeandpart}&84.5& 61.2& 88.4& 66.7& 96.1& 98.3& 85.7& 74.7& 79.5& 69.1& 80.4 \\
 Action Mask~\cite{mini}&82.3& 69.2& 91.1& 67.3& 91.5& 96.0& 84.4& 71.2& 90.5& 60.6& 80.4 \\
 ResNet-50~\cite{He2015}&88.9& 75.2& 90.7& 75.9& 95.8& 97.5& 81.1& 69.5& 84.9& 53.9& 81.4 \\

 \hline
ours &89.2& 79.0& 92.2& 78.7& 95.9 &97.2 &87.1& 73.1 &86.0 &61.9& \textbf{84.0}\\
\hline

\end{tabular}
\end{table*}

\begin{table*}[tbp]
\caption{\label{voctest}Preformance (mAP) on the PASCAL VOC 2012 Action test set}
\footnotesize
\centering
\begin{tabular}{|c|cccccccccc|c|}
\hline
\multirow{2}{*}{method}  & \multirow{2}{*}{jumping}&\multirow{2}{*}{phoning} &
playing &\multirow{2}{*}{reading}&
riding&riding&\multirow{2}{*}{running}&
taking&using&\multirow{2}{*}{walking}&\multirow{2}{*}{mAP}\\
&&&instrument&&bike&horse&&photo&computer&&\\
 \hline
 Whole\&Parts~\cite{wholeandpart}&84.7& 67.8& 91.0& 66.6& 96.6& 97.2& 90.2& 76.0& 83.4& 71.6& 82.6 \\
 VGG-16~\cite{vgg}&- &-& -& - &-& -& -& - &-& - &79.2 \\
 
 ResNet-50~\cite{He2015}&91.1& 76.1& 89.3& 72.4& 97.3& 97.2& 81.8& 70.2& 84.0& 68.1& 82.7 \\
 Action Mask~\cite{mini}&83.5& 70.6& 92.3& 68.7& 94.8& 96.7& 87.5& 70.7& 86.3& 64.6& 81.6 \\
 \hline
ours &93.0& 81.4& 92.6& 77.6& 97.7 &97.8 &87.6& 74.3 &87.1 &75.1& \textbf{86.4}\\
\hline

\end{tabular}
\end{table*}

\begin{table}[tbp]
\caption{\label{stanford}Preformance (mAP) on the Stanford-40 dataset}
\centering
\begin{tabular}{|c|c|}
\hline
method  & mAP\\
\hline
TDP~\cite{topdown}  & 80.6\\
ResNet-50~\cite{He2015}  & 80.2\\
Action Mask~\cite{mini}  & 80.8\\
\hline
ours &  \textbf{83.4}\\
\hline
\end{tabular}
\end{table}

\begin{figure*}[tbp]
\centering
\includegraphics[width=0.9\linewidth]{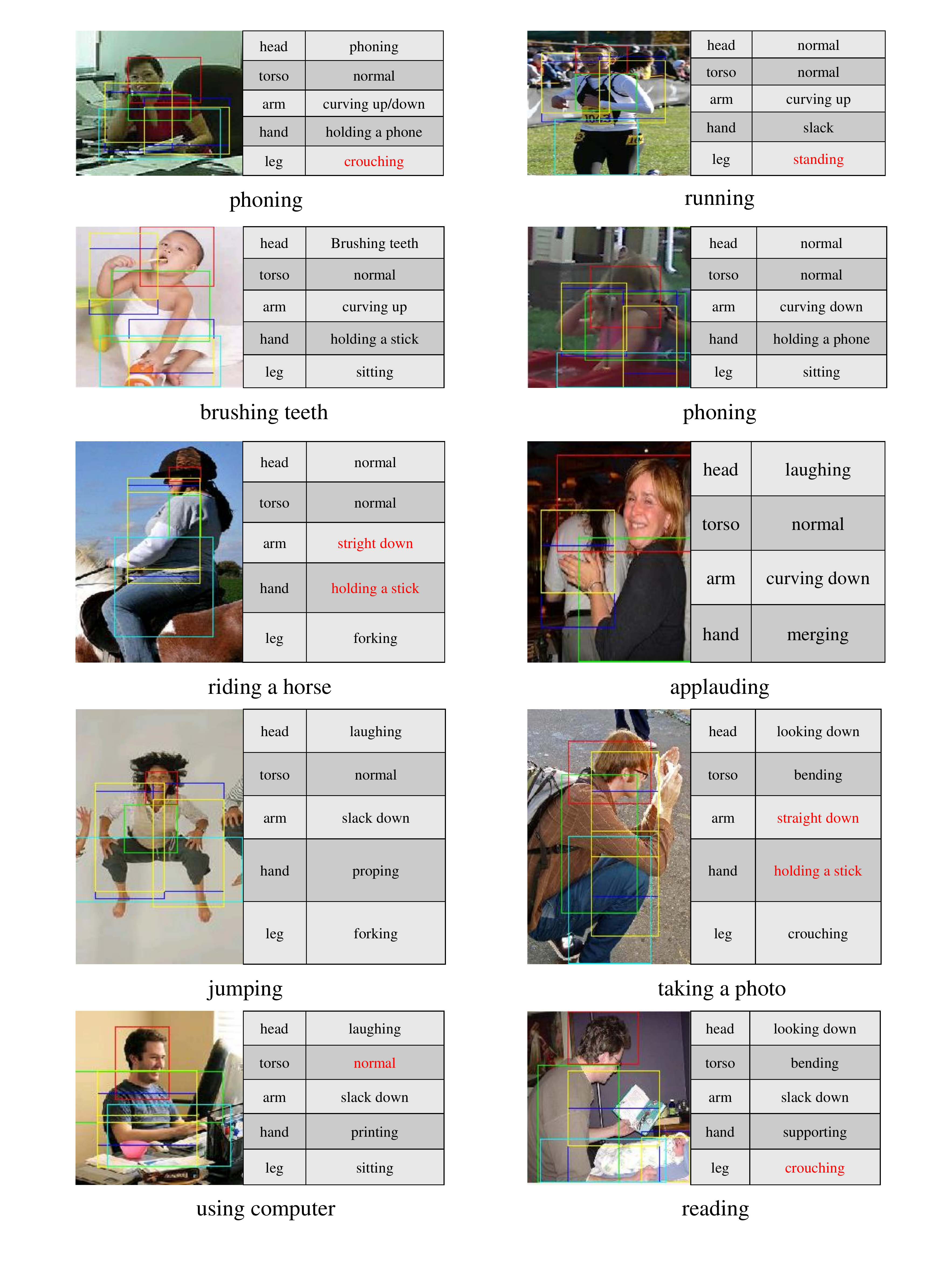}
\caption{Visualization of part localization and semantic body part actions and final action prediction on PASCAL VOC 2012 test set and Stanford-40 dataset. 
Head, torso, arms, hands and legs are shown in red, green, blue, yellow and cyan, respectively.
}
\label{vis}
\end{figure*}

\subsection{Comparison with existing methods}

We compare our approach with the state-of-the-art methods on the two datasets.
Tab.\ref{vocval} reports the results on PASCAL VOC 2012 Action validation dataset~\cite{pascalvoc2012}, the results on test set are shown in Tab.\ref{voctest}.
VGG-16~\cite{vgg} and ResNet-50~\cite{He2015} simply take CNN as feature extractor, and classify fc layer features by a linear SVM. ResNet-50 are essentially our baseline.
Gkioxari \etal~\cite{wholeandpart} use deep poselets to detect head,
torso and legs regions and concatenate the corresponding features.
Gkioxari \etal~\cite{contextualaction} view the bounding box as the first region, and search a secondary region in the whole image with overlap constrained. 
Zhang \etal~\cite{mini} propose a method that accurately delineate the foreground regions of underlying human-object interactions. The foreground region is also called ``action mask".

Defer from the above mentioned methods, our proposed method dose not employ context information. We only use images within the bounding boxes and focus on the part actions.
Our method outperforms all of the methods that use no context information significantly and achieves a gain over the second best result by 3.7\%. 
In categories that part actions are more important, such as ``jumping" and ``walking", our method ps promising results.
The results utilizing context is reported in supplementary material.

Tab.\ref{stanford} shows the comparison on Stanford-40 dataset~\cite{Yao12}. Our method outperforms the state-of-the-art by 2.6\%, compared with the methods that use no context information .

\subsection{Visual results}

Fig.\ref{vis} shows visual results of part localization, semantic part action classification and final action prediction of the proposed method. Head, torso, arms, hands and legs are shown in red, green, blue, yellow and cyan correspondingly.
For most instances, part actions provide informative cues. See the example of ``running'', curving arms defined in our method help distinguish this action from ``walking''. Based on ``printing'' hand we can infer that the man of row $5$ and column $1$ is using computer, his laughing head also help to distinguish this action from ``reading''.

\section{Conclusion}\label{conclusion}

This paper proposes single image action recognition algorithm which is based on the idea of semantic body part actions.
It is based on the observation that the human action is a combination of meaningful body part actions. 
Human body is divided into five parts: head, torso, arms, hands and legs. And for each of the body parts, semantic body part actions is defined. 
A deep neural network based system is proposed: first, body parts are localize by a Semi-FCN network. Second, for each body parts, a Part Action Res-Net is used to predict semantic body part actions. And final, we use SVM to combine the body part actions and predict the entire body action.
Experiments on two dataset: PASCAL VOC 2012 and Stanford-40 reports accuracy improvement from state-of-the-art by 3.8\% and 2.6\% respectively.

{\small
\bibliographystyle{ieee}
\bibliography{egbib}

\begin{thebibliography}{10}\itemsep=-1pt

\bibitem{andriluka14cvpr}
M.~Andriluka, L.~Pishchulin, P.~Gehler, and B.~Schiele.
\newblock 2d human pose estimation: New benchmark and state of the art
  analysis.
\newblock In {\em CVPR}, 2014.

\bibitem{poselet}
L.~Bourdev and J.~Malik.
\newblock Poselets: Body part detectors trained using 3d human pose
  annotations.
\newblock In {\em ICCV}, 2009.

\bibitem{pascalpart}
X.~Chen, R.~Mottaghi, X.~Liu, S.~Fidler, R.~Urtasun, and A.~Yuille.
\newblock Detect what you can: Detecting and representing objects using
  holistic models and body parts.
\newblock In {\em CVPR}, 2014.

\bibitem{hoi}
P.-Y.~P. Chi, Y.~Li, and B.~Hartmann.
\newblock Enhancing cross-device interaction scripting with interactive
  illustrations.
\newblock In {\em HFCS}, 2016.

\bibitem{imagenet}
J.~Deng, R.~Socher, L.~Fei-Fei, W.~Dong, K.~Li, and L.-J. Li.
\newblock Imagenet: A large-scale hierarchical image database.
\newblock In {\em CVPR}, 2009.

\bibitem{phrasetree}
C.~Desai and D.~Ramanan.
\newblock Detecting actions, poses, and objects with relational phraselets.
\newblock In {\em ECCV}, 2012.

\bibitem{pascalvoc2012}
M.~Everingham, L.~V. Gool, C.~Williams, J.~Winn, and A.~Zisserman.
\newblock The pascal visual object classes challenge 2012 (voc2012) results.
\newblock 2012.

\bibitem{wholeandpart}
G.~Gkioxari, R.~Girshick, and J.~Malik.
\newblock Actions and attributes from wholes and parts.
\newblock In {\em ICCV}, 2015.

\bibitem{contextualaction}
G.~Gkioxari, R.~Girshick, and J.~Malik.
\newblock Contextual action recognition with r*cnn.
\newblock In {\em ICCV}, 2015.

\bibitem{He2015}
K.~He, X.~Zhang, S.~Ren, and J.~Sun.
\newblock Deep residual learning for image recognition.
\newblock {\em CVPR}, 2016.

\bibitem{SMSP}
F.~S. Khan, J.~van~de Weijer, R.~M. Anwer, M.~Felsberg, and C.~Gatta.
\newblock Semantic pyramids for gender and action recognition.
\newblock {\em TIP}, 23(8):3633--3645, 2014.

\bibitem{fcn}
J.~Long, E.~Shelhamer, and T.~Darrell.
\newblock Fully convolutional networks for semantic segmentation.
\newblock In {\em CVPR}, 2015.

\bibitem{pascalvoc2010}
M.Everingham, L.~Gool, C.Williams, J.Winn, and A.Zisserman.
\newblock The pascal visual object classes challenge 2010 (voc2010) results.
\newblock 2010.

\bibitem{vgg}
K.~Simonyan and A.~Zisserman.
\newblock Very deep convolutional networks for large-scale image recognition.
\newblock In {\em ICLR}, 2015.

\bibitem{c3d}
D.~Tran, L.~Bourdev, R.~Fergus, L.~Torresani, and M.~Paluri.
\newblock C3d: Generic features for video analysis.
\newblock In {\em ICCV}, 2015.

\bibitem{selectivesearch}
J.~R.~R. Uijlings, K.~E.~A. van~de Sande, T.~Gevers, and A.~M. Smeulders.
\newblock Selective search for object recognition.
\newblock In {\em IJCV}, 2013.

\bibitem{VlFeat}
A.~Vedaldi and B.~Fulkerson.
\newblock Vlfeat -- an open and portable library of computer vision algorithms.
\newblock In {\em Proceedings of the 18th annual {ACM} international conference
  on Multimedia}, 2010.

\bibitem{onevsall}
X.~Z.~H. Xiong, W.~Zhou, and Q.~Tian.
\newblock Fused one-vs-all features with semantic alignments for fine-grained
  visual categorization.
\newblock {\em IEEE Transactions on Image Processing}, 25:878--892, 2016.

\bibitem{cap1}
K.~Xu, J.~Ba, R.~Kiros, K.~Cho, A.~Courville, R.~Salakhutdinov, R.~Zemel, and
  Y.~Bengio.
\newblock Show, attend and tell: Neural image caption generation with visual
  attention.
\newblock In {\em arXiv:1502.03044}, 2015.

\bibitem{25D}
B.~Yao and L.~Fei-Fei.
\newblock Action recognition with exemplar based 2.5d graph matching.
\newblock In {\em ECCV}, 2012.

\bibitem{Yao12}
B.~Yao, X.~Jiang, A.~Khosla, A.~L. Lin, L.~Guibas, and L.~Fei-Fei.
\newblock Human action recognition by learning bases of action attributes and
  parts.
\newblock In {\em ICCV}, 2011.

\bibitem{Yao11}
B.~Yao, A.~Khosla, and L.~Fei-Fei.
\newblock Combining randomization and discrimination for fine-grained image
  categorization.
\newblock In {\em Proceedings of the International Conference on Computer
  Vision and Pattern Recognition}, 2011.

\bibitem{parteccv}
N.~Zhang, J.~Donahue, R.~Girshick, and T.~Darrell.
\newblock Part-based r-cnns for fine-grained category detection.
\newblock In {\em ECCV}, 2014.

\bibitem{finefcn}
N.~Zhang, E.~Shelhamer, Y.~Gao, and T.~Darrell.
\newblock Fine-grained pose prediction, normalization, and recognition.
\newblock In {\em arXiv:1511.07063}, 2015.

\bibitem{mini}
Y.~Zhang, L.~Cheng, J.~Wu, J.~Cai, M.~N. Do, and J.~Lu.
\newblock Action recognition in still images with minimum annotation efforts.
\newblock {\em IEEE Transactions on Image Processing}, 25:5479--5490, 2016.

\bibitem{topdown}
Z.~Zhao, H.~Ma, and X.~Chen.
\newblock Semantic parts based top-down pyramid for action recognition.
\newblock {\em Pattern Recognition Letters}, 84:134--141, 2016.

\end{thebibliography}
}

\end{document}